\documentclass[10pt,twocolumn,letterpaper]{article}

\usepackage[pagenumbers]{wacv} % To force page numbers, e.g. for an arXiv version

\usepackage[accsupp]{axessibility}  % Improves PDF readability for those with disabilities.
\usepackage[dvipsnames]{xcolor}
\usepackage{amsmath}
\usepackage{amssymb}
\usepackage{subcaption}
\usepackage{booktabs}
\usepackage{tikz}
\usetikzlibrary{spy}

\usepackage{xspace}
\def\etal{et al.\xspace}
\def\eg{e.g.\xspace}
\def\ie{i.e.\xspace}

\usepackage[normalem]{ulem}

\newcommand{\norm}[1]{\left\Vert#1\right\Vert}

\let\originalleft\left
\let\originalright\right
\renewcommand{\left}{\mathopen{}\mathclose\bgroup\originalleft}
\renewcommand{\right}{\aftergroup\egroup\originalright}

\definecolor{best}{rgb}{1, 0.7, 0.7}
\definecolor{second}{rgb}{1, 0.85, 0.7}
\definecolor{tab1}{rgb}{0.31,0.47,0.65}
\definecolor{tab2}{rgb}{0.95,0.56,0.17}

\newcommand{\imagewithzoom}[8]{%
    \begin{tikzpicture}[spy using outlines={lens={scale=#7}, size=#6}]
        \node[draw=black, line width=2pt, inner sep=0pt] at (0, 0)  
            {\includegraphics[width=\textwidth, keepaspectratio, trim={#8}, clip]{#1}};
        \spy [red] on (#2,#3) in node at (#4,#5);
    \end{tikzpicture}%
}

\usepackage{pifont}

\usepackage[normalem]{ulem}
\definecolor{iccvblue}{rgb}{0.21,0.49,0.74}

\usepackage{fancyhdr}
\usepackage{mdframed}

\fancypagestyle{firstpage}{
  \fancyhf{}

  \fancyfoot[C]{%
    \begin{minipage}{0.97\linewidth}
    \vspace{8pt}
    \begin{mdframed}[linewidth=0.5pt, roundcorner=4pt]
        \footnotesize
        © 2025 IEEE. Personal use of this material is permitted. Permission from IEEE must be obtained
        for all other uses, including reprinting, republishing, or redistribution. This is the authors'
        accepted manuscript of the paper: L. Holland et al., “Transformer-Based Inpainting for Real-Time 3D Streaming in Sparse
        Multi-Camera Setups,” in \textit{IEEE/CVF Winter Conference on Applications of Computer Vision (WACV)}, 2026.\\
        The final version will be available via IEEE Xplore once published.
    \end{mdframed}
    \end{minipage}
  }
}
\definecolor{wacvblue}{rgb}{0.21,0.49,0.74}
\usepackage[pagebackref,breaklinks,colorlinks,allcolors=wacvblue]{hyperref}
\def\wacvPaperID{954} % *** Enter the WACV Paper ID here
\def\confName{WACV}
\def\confYear{2026}

\title{Transformer-Based Inpainting for Real-Time 3D Streaming\\in Sparse Multi-Camera Setups}

\author{Leif Van Holland
\and
Domenic Zingsheim
\and
Mana Takhsha
\and
Hannah Dröge
\and
Patrick Stotko
\and
Markus Plack
\and
Reinhard Klein
\and
\vspace{-0.3cm}
\\
\vspace{0.0cm}
{\tt\small \{holland,zingsheim,takhsha,droege,stotko,mplack,rk\}@cs.uni-bonn.de}\\
University of Bonn, Germany
}

\begin{document}
\maketitle
\thispagestyle{firstpage}
\begin{abstract}
High-quality 3D streaming from multiple cameras is crucial for immersive experiences in many AR/VR applications.
The limited number of views - often due to real-time constraints - leads to missing information and incomplete surfaces in the rendered images.
Existing approaches typically rely on simple heuristics for the hole filling, which can result in inconsistencies or visual artifacts.
We propose to complete the missing textures using a novel, application-targeted inpainting method independent of the underlying representation as an image-based post-processing step after the novel view rendering.
The method is designed as a standalone module compatible with any calibrated multi-camera system.
For this we introduce a multi-view aware, transformer-based network architecture using spatio-temporal embeddings to ensure consistency across frames while preserving fine details.
Additionally, our resolution-independent design allows adaptation to different camera setups, while an adaptive patch selection strategy balances inference speed and quality, allowing real-time performance.
We evaluate our approach against state-of-the-art inpainting techniques under the same real-time constraints and demonstrate that our model achieves the best trade-off between quality and speed, outperforming competitors in both image and video-based metrics.

\end{abstract}    
\section{Introduction}
\label{sec:intro}

\begin{figure}[t]
    \centering
    \includegraphics[width=\linewidth]{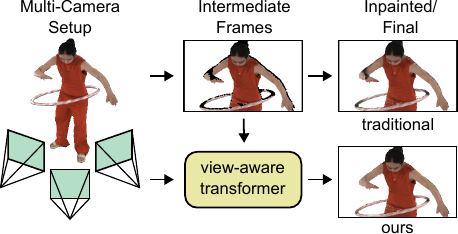}
    \caption{
      Streamed content from a multi-camera setup (left)
      is prone to incomplete textures (center) because of missing information in the sparse viewpoints.
      To fix this, we propose a transformer-based inpainting method that efficiently incorporates information from the original images and thus surpasses traditional inpainting on the reconstruction alone (right).
    }
    \label{fig:teaser}
    \vspace{-12pt}
\end{figure}

Template-free 3D streaming holds immense potential across various domains, including entertainment (e.g. sports, arts, concerts, and film), telepresence, and medical applications,
but remains challenging, especially when targeting consumer hardware and AR/VR devices~\cite{xu20244k4d, dou2017motion2fusion,dou2016fusion4d,zhang2022virtualcube,dai2024real}.
The enormous amounts of data produced by multi-camera setups are tricky to handle in real-time applications, pushing the need for careful selection and distillation of the underlying information~\cite{carballeira2021fvv}.
This directly contradicts the common notion in multi-view reconstruction that sparse viewpoints lead to reproductions of the scene that often contain incomplete geometry or textures  \cite{yu2020point, oh2010virtual}.
Those unseen regions are a fundamental challenge to all such methods, and their effects drastically reduce the perceptual quality of the video stream 
(see \Cref{fig:teaser}), highlighting the need for sophisticated methods to fill the gaps.

While video inpainting methods offer a straightforward solution to this problem, state-of-the-art methods are not designed for such a use case, with but a few approaches being capable of (near) real-time performance \cite{zeng2020learning, Liu_2021_DSTT}.
Instead, a major focus of recent works \cite{zhang2022flow, zeng2020learning, gu2024advanced} has been the temporal consistency needed to produce perceptually good results and the efficient feature propagation needed to access the information from other frames, potentially bridging long temporal and spatial gaps.
This is different from our use case, where the limited information contained in the incomplete novel view makes it difficult to generate plausible content.
There is even a high likelihood that the necessary information is not contained in any of the past frames, as they were produced by the same setup.
Improving the feature propagation beyond the single view is therefore a central point of optimization, as it reduces the burden of the content hallucination task, which is highly ill-posed.

Unlike 3D inpainting methods that directly complete geometry or radiance fields \cite{wang2024gscream,lin2024maldnerf,wu2025aurafusion, huang20253d}, our work targets 2D video inpainting of rendered views within a real-time 3D streaming pipeline, where available multi-view information has already been fused by a geometry proxy.
We argue that the original images used to generate the 3D representation offer rich information for a model to use during the inpainting process, much of which is not contained in the novel view.
We propose a learning-based method that leverages this readily available data from the given input views using a transformer-based architecture, which is naturally well-suited for this task.
Our method operates on feature-space patches from both the target image and the context images that are the original camera views as well as past frames from all views.
To facilitate information transfer, we introduce a spatio-temporal encoding for the context patches and their relative coordinates, utilizing the underlying 3D proxy to make better use of the contextual information.
For faster inference, we propose a top-k filtering mechanism and demonstrate real-time performance with a negligible loss in quality.
We evaluate our method against state-of-the-art inpainting approaches on real-world data demonstrating superior performance across image and video metrics, and study the impact of our model components as well as the performance-runtime tradeoff of our speed-up strategy.
We focus on human-centric, foreground-matted sequences typical of telepresence pipelines.

In summary, our main contributions are as follows.

\begin{itemize}
    \item We introduce a novel, multi-view aware transformer-based inpainting network for real-time video inpainting as a general post-processing step in 3D streaming pipelines.
    \item We propose a spatio-temporal embedding that enhances feature propagation of multi-view information using a geometry proxy for reprojection.
    \item We design a patch filtering based on spatio-temporal locality to adjust the amount of patches required during inference, allowing a trade-off between speed and accuracy.
\end{itemize}

The source code of our implementation is available at \url{https://github.com/vc-bonn/transformer-based-inpainting}.

\section{Related work}
\label{sec:related_work}

\begin{figure*}[t]
    \centering
    \includegraphics[width=\linewidth]{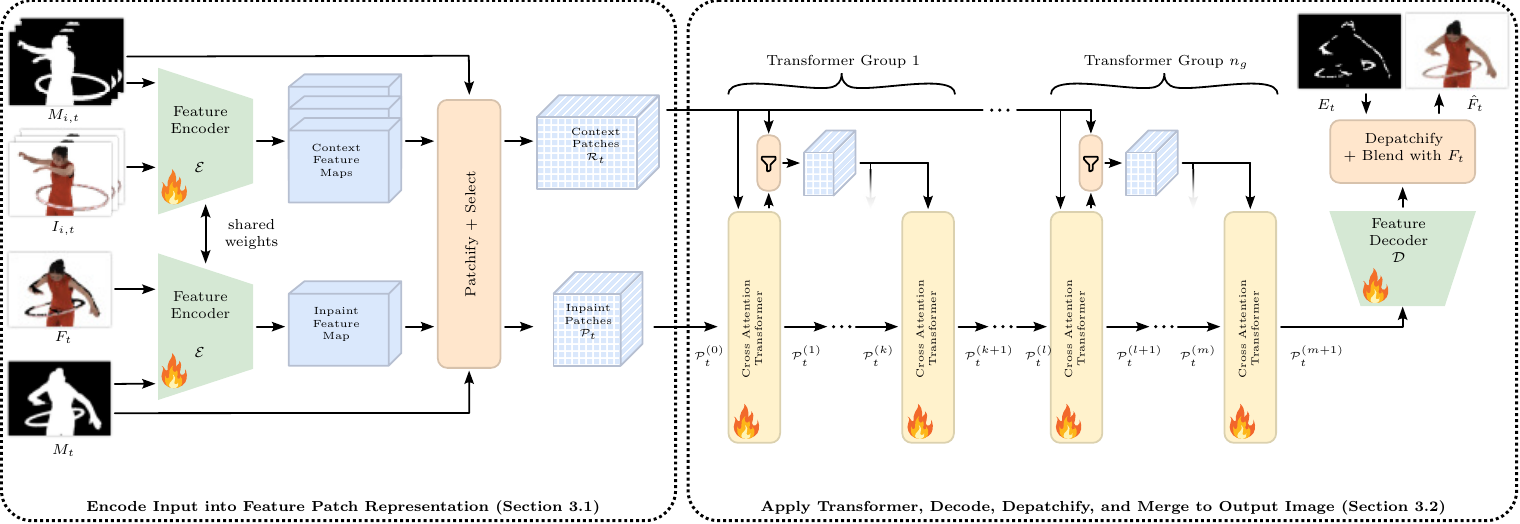}
    \captionof{figure}{
    Overview of the proposed transformer-based inpainting pipeline. The framework consists of two main stages: (1) \textit{Feature Encoding}, where input and context images are encoded into feature representations and split into patches equipped with their spatio-temporal coordinates, (2) \textit{Context Aggregation and Decoding}, utilizing a series of transformer groups and the contextual information to update the inpaint patches, that are finally decoded and blended with the known image regions. The flame symbol indicates a module with trainable parameters.}
    \vspace{-12pt}
    \label{fig:architecture}
\end{figure*}

In the following section, we review relevant work on image and video inpainting, followed by a discussion of shape and texture completion methods, which predominantly use inpainting techniques.

\subsection{Image Inpainting}
Image inpainting methods have been explored for many years, progressing from traditional interpolation \cite{getreuer2012total} and patch-based techniques \cite{wong2008nonlocal} to advanced deep learning methods \cite{corneanu2024latentpaint, chen2024dnnam, chen2024image}. 
Among these modern techniques, encoder-decoder architectures have emerged as a popular choice for reconstructing missing regions in images \cite{liu2020rethinking, wang2018image, zeng2019learning}, while multi-stage learning strategies further improved performance \cite{nazeri2019edgeconnect, xiong2019foreground}.
% transformers
In recent years, attention mechanisms and vision transformers have been integrated into image inpainting \cite{cao2022learning, deng2022hourglass}.  One of the pioneering approaches using 
contextual attention has been introduced by Yu \etal \cite{yu2018generative}, which has since been widely adopted and refined \cite{liu2019coherent, li2022mat, chen2023dgca}.
For instance, Liu \etal \cite{liu2019coherent} introduced a semantic attention layer that incorporates semantic relevance, while Qin \etal \cite{qin2021multi} employed multi-scale attention to capture both semantics and details. Similarly,  
Wang \etal \cite{wang2019musical} use multi-scale attention to take advantage of background information from the given image.
%
% Generative - GAN DIffusion
Generative approaches have also gained significance in image inpainting.
 Earlier methods  primarily  relied on Generative Adversarial Networks (GANs) to inpaint missing regions \cite{liu2021pd, zhao2021large, zheng2022image}, whereas more recently, diffusion models became a powerful tool for improved restoration capabilities \cite{ju2024brushnet, liu2024structure, xie2023smartbrush, kim2025rad, wang2025guidpaint}.
%
%Pluralistic Inpainting
Given that image inpainting is inherently ill-posed, with no single correct solution, research has increasingly shifted towards pluralistic methods. These approaches generate multiple plausible inpainted outcomes, addressing the task's inherent ambiguity \cite{zheng2019pluralistic, zhao2020uctgan}.
For a more detailed overview, refer to \cite{quan2024deep, zhang2023image}.

While most image inpainting methods operate on single images, our work builds on transformer architectures to address multi-view settings with geometric consistency and temporal video inpainting.

\subsection{Video Inpainting}
Early methods formulate video inpainting as patch-based optimization \cite{huang2016temporally, newson2014video}, which struggles with complex motion.
With deep learning, approaches based on 3D convolutions \cite{wang2019video, chang2019free, hu2020proposal} and temporal-shift mechanisms  \cite{lin2019tsm, zou2021progressive} improved temporal coherence but still face challenges in modeling motion explicitly, motivating flow-guided methods \cite{kim2019deep, chang2019vornet, zhang2019internal, li2022towards}.
Examples include flow-guided pixel propagation \cite{xu2019deep}, handling occluded objects \cite{ke2021occlusion}, motion-edge guidance to avoid over-smoothing \cite{gao2020flow}, and feature-level propagation for speed \cite{li2022towards}.

Transformers extend temporal context and long-range dependencies for inpainting \cite{li2022mat, wan2021high, chen2021deep, Liu_2021_FuseFormer, sun2025mask}, including flow-aware variants \cite{zhang2022flow, zhou2023propainter, li2025aggregating}. Architectures such as DSTT \cite{Liu_2021_DSTT}, FuseFormer \cite{Liu_2021_FuseFormer}, and E2FGVI \cite{li2022towards} deliver strong quality, and ProPainter further improves propagation/transformer design \cite{zhou2023propainter}. Diffusion-based models advance image/video inpainting quality \cite{anciukevivcius2023renderdiffusion, li2025diffueraser, liu2025eraserdit, lee2025video}, sometimes guided by optical flow or text  \cite{gu2024advanced,zhang2024avid, zi2025cococo}, but typically incur substantial computational cost that limits real-time use. Overall, this line of work is primarily designed for offline processing with access to future frames.

To support streaming scenarios, Thiry \etal \cite{thiry2024towards} introduced online variants of DSTT, FuseFormer, and E2FGVI by conditioning only on past frames. These models are the closest prior approaches available for online video inpainting; accordingly, we adopt them as baselines.
Our method is most related to transformer-based inpainting (\eg, DSTT, FuseFormer), but differs in that it explicitly leverages multi-view geometry via reprojection-aware spatio-temporal embeddings to aggregate context across views and time under real-time constraints in a sparse multi-camera streaming setup. Unlike offline methods, our formulation targets online inference and a resolution-independent design.

\subsection{Shape and Texture Completion}
Several works address completion in 3D representations. Classical hole filling has been studied in depth- or view-synthesis contexts \cite{oh2010virtual, zou2014view,liu2023novel, luo2019disocclusion}, while geometry completion methods aim to reconstruct missing regions in meshes, point clouds, or volumetric representations \cite{lim2016learning}. 
Another line of research focuses on hole-filling in 3D geometry. In this context, learning-based methods have been explored, for instance by using generative adversarial networks \cite{yu2020point}, adaptations of 2D inpainting models for 3D completion \cite{hernandez20243d}, or point-based networks \cite{sipiran2022data}.
More recent approaches such as NeRFiller \cite{weber2024nerfiller}, Gscream \cite{wang2024gscream}, MALD-NeRF \cite{lin2024maldnerf}, AuraFusion360 \cite{wu2025aurafusion}, and 3DGIC \cite{huang20253d} pursue generative scene completion in neural radiance fields or Gaussian splats, explicitly enforcing cross-view consistency in 3D.

In contrast, our setting differs fundamentally: we do not inpaint the 3D representation itself, but rather perform 2D video inpainting within a 3D streaming pipeline. Specifically, our method takes multi-view fused renderings (\eg, RIFTCast~\cite{zingsheim2025riftcast}) and fills residual occlusions and missing regions in these images under strict real-time constraints. This task is complementary to 3D completion approaches and more closely tied to telepresence applications, where low-latency corrections of rendered frames are critical.

\section{Method}

Our method imposes minimal constraints on the underlying 3D streaming approach, ensuring compatibility with a wide range of real-time frameworks.
At each time step $ t $, we assume the reconstruction algorithm provides a geometric representation $\mathcal{G}_t$ along with the input RGB images $I_{i,t}$ captured by cameras $i \in \{1,..,N\}$.
Using this data, we can synthesize novel views $F_t$.
However, due to real-time constraints, these novel views might contain inaccuracies.
The goal of our inpainting model $\mathcal{T}$ is to correct these errors and infer complete output frames $\hat{F}_t$ by leveraging both the available data and information from previous frames $\tau \leq t$, which we refer to as the \emph{context input}:
\begin{equation}
    \hat{F}_t = \mathcal{T}\left(F_t \, | \, \mathcal{G}_t, \{I_{i,\tau}\}_{i\leq N, \tau\leq t} \right)
\end{equation}
Our inpainting model consists of three primary components: an image encoder $\mathcal{E}$, groups of transformer blocks operating on feature patches, and a decoder $\mathcal{D}$, as illustrated in \Cref{fig:architecture}.
Additionally, we assume the availability of foreground masks $M_{i,t}$ for each input view $I_{i,t}$ and a corresponding mask $M_t$ for the target view.
These masks are typically accessible in most systems.
We also utilize an error map $E_t$, which identifies regions of the output frame requiring inpainting.
For the error maps $E_t$, we assume that some error detection is available, \eg, tracking occluded pixels during the reconstruction process.

\subsection{Encoding and Patch Extraction}
\label{sec:feature_encoding}

The CNN image encoder $\mathcal{E}$ based on FuseFormer~\cite{Liu_2021_FuseFormer} processes both the context input $\{I_{i,t},M_{i,t}\}$ and the novel‐view input $\{F_{t},M_{t}\}$ independently, 
transforming them into hierarchical, high-dimensional feature representations taken from multiple stages in the CNN.
From the available geometry proxy $\mathcal{G}_t$, we can optionally generate pseudo depth information by re-rendering $\mathcal{G}_t$ using the known camera parameters of the calibrated capturing system and use these pseudo‐depth maps as an auxiliary input passed as an extra channel along with the context and novel‐view images.
We then repeat this encoding process for all subsequent context frames, generating a set of feature representations.
In addition, we follow Thiry \etal~\cite{thiry2024towards} and encode context frames from timesteps adjacent to the current frame: the $n_c$ frames immediately preceding the current frame $t$, ${\{\,t - j \mid j = 1, 2, \dots, n_c\}}$, and the $n_w$ frames further in the past with spacing factor ${k_w > 1}$ to cover a wider context without considering every past frame, $\{{\,t - k_w\,j \mid j = 1, 2, \dots, n_w\}}$.

\begin{figure}[t]
    \centering
    \begin{subfigure}{0.43\linewidth}
        \includegraphics[width=\linewidth]{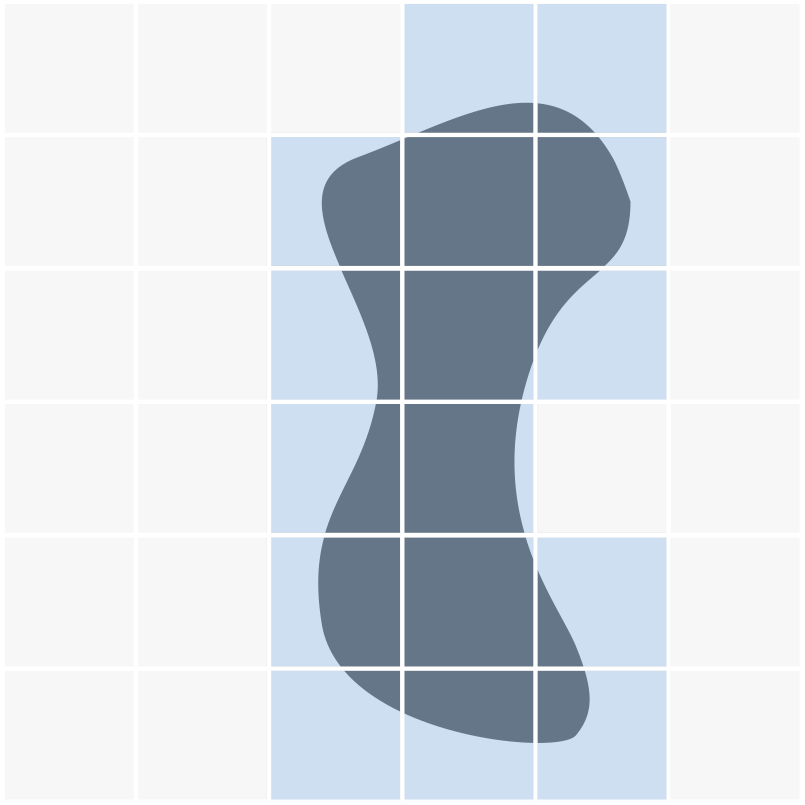}
        \caption{Context Feature Map}
    \end{subfigure}
    \hfill
    \begin{subfigure}{0.43\linewidth}
        \includegraphics[width=\linewidth]{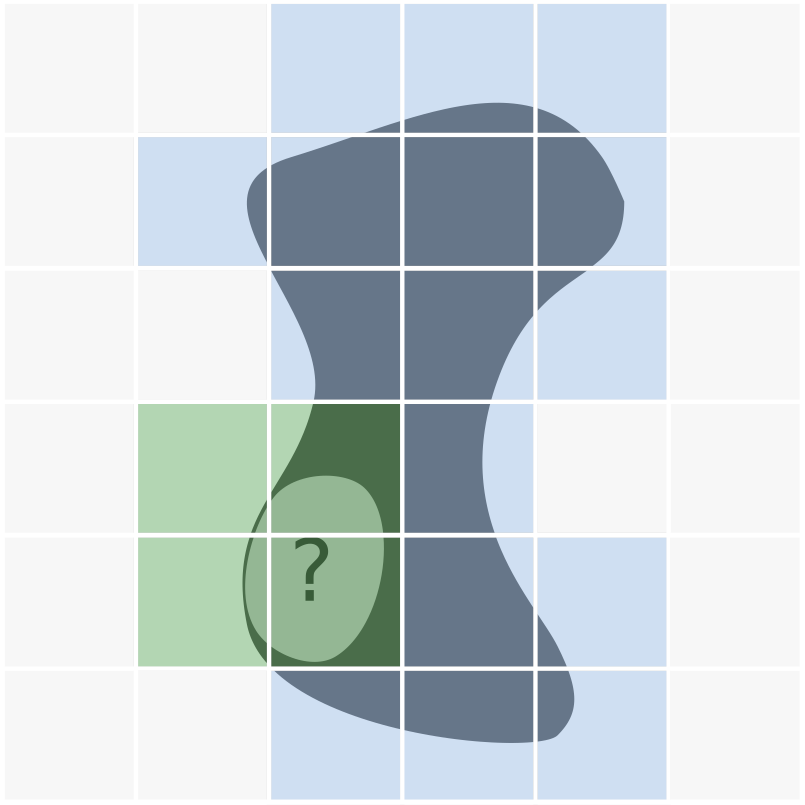}
        \caption{Inpaint Feature Map}
    \end{subfigure}
    \caption{Feature maps are split into overlapping patches: background‐only patches are pruned, and object patches are kept as context $\mathcal{R}_t$ (blue). In the inpaint feature map, patches with missing pixels (green) constitute $\mathcal{P}_t$, while patches without missing pixels are added to the context. In this illustration, we show patches as non‐overlapping for clarity, even though they are overlapping in practice.}
    %{}
    \vspace{-12pt}
    \label{fig:context_patch}
\end{figure}

The generated feature maps are divided into small, overlapping patches, and any patch consisting entirely of background is discarded (see \Cref{fig:context_patch}).
From the inpaint feature map, patches that contain no pixels requiring inpainting are also added to the context set instead. 
The remaining inpaint patches (with missing pixels) form the input set $\mathcal{P}_t$, while all retained patches form the context set $\mathcal{R}_t$. 

Each patch ${p \in \mathcal{P}_t}$ or ${r \in \mathcal{R}_t}$ is associated with spatiotemporal coordinates ${x_p, x_r \in \mathbb{R}^3}$, enabling the model to know exactly where (and when) each patch is located.
Concretely, each vector ${x\in\mathbb{R}^3}$ encodes the screen-space coordinates of the patch center (normalized to $[0,1]$) together with the timestep at which the patch appears, which is also normalized using the maximal window-size $k_w \, n_w$. 
To relate context patches to the novel view, we project
\begin{equation}
   \hat{x}_r = C_{\mathcal{G}_t}\bigl(x_r\bigr),
   \label{eq:spatiotemporal_coords}
\end{equation}
where $C_{\mathcal{G}_t}(\cdot)$ is the reprojection function that maps a screen-space coordinate from the context camera’s view into the target (novel) camera’s view.

\subsection{Transformer Blocks and Decoding}
\label{sec:transformer_blocks}

The input patches $\mathcal{P}_t$ are processed by a series of $n_g$ transformer groups, each comprising  $n_b$ blocks. Within each block, the patch sequence is updated by attending to the context patch sequence $\mathcal{R}_t$.
At the $k$-th block, the normalized input $\mathcal{P}_t^{(k-1)}$ is updated as
\begin{equation}
    \mathcal{P}_{t}^{(k)} = \mathcal{P}_{t}^{(k-1)} + A\left(W_Q^{(k)} \, \mathcal{P}_t^{(k-1)}, \, W_K^{(k)} \, \mathcal{R}_t, \, W_V^{(k)} \, \mathcal{R}_t \right),
\end{equation}
where  ${\mathcal{P}_t^{(0)}=\mathcal{P}_t}$, and  $W_Q^{(k)}, W_K^{(k)}, W_V^{(k)}$ are learned projection operators of the $k$-th block, and the result is then passed through a standard feed-forward block.
Here, we incorporate each patch’s spatiotemporal coordinates via a decomposed 3D variant of rotary positional embeddings (RoPE) \cite{su2024roformer,heo2024rotary}  in the attention computation. Thus,
\begin{equation}
    A(Q,K,V)=\sigma\left(\frac{\text{RoPE}(Q, x_Q)\cdot \text{RoPE}(K, \hat{x}_K)^T}{\sqrt{D}}\right) V,
\end{equation}
where $x_Q$ and $\hat{x}_K$ are the (projected) 3D spatiotemporal coordinates associated with the query and key patches, respectively.
$D$ represents the size of the feature dimension and $\sigma$ denotes the softmax function.
For a pair of patches, RoPE encodes a relative position across multiple frequencies without explicitly computing the pairwise coordinate distances.
For more details, refer to \cite{su2024roformer}.

It can be expected that many of the context patches do not contain valuable information for the inpainting.
We therefore apply patch sparsification right after the very first transformer within each group
to improve runtime performance by retaining only the most relevant patches.
We compute the sum of attention weights for every token in the context patches and keep only the top-k tokens according to that sum and train the mask with a straight-through estimator~\cite{bengio2013estimating,shazeer2017outrageously}, following the token-pruning approach of Cordonnier~\etal~\cite{cordonnier2021differentiable}.
Once all transformer groups have processed the patches, the resulting patch representations are forwarded to a deconvolutional decoder network $\mathcal{D}$ that computes an RGB patch independently from each feature patch.
Next, the reconstructed patches are reinserted into their original location by linearly blending overlapping pixels,
resulting in an intermediate RGB image~$\tilde{F}_t$, that is linearly blended with the input using the error mask $E_t$ yielding the final output
\begin{equation}
    \hat{F}_t = E_t \odot \tilde{F}_t + (1 - E_t) \odot {F}_t,
\end{equation}
where $ \odot$ denotes elementwise multiplication.

\subsection{Loss and Efficient Inference}
The model is then trained with the combination of an $\ell_1$ image loss and an adversarial loss,
\begin{equation}
    \mathcal{L} = \lambda_\text{img} \, (\mathcal{L}_{\text{in}} + \mathcal{L}_{\text{out}})  + \lambda_\text{adv} \, \mathcal{L}_{\text{adv}},
\end{equation}
where $\lambda_{\mathrm{img}}$ and $\lambda_{\mathrm{adv}}$ are weighting coefficients controlling the trade‐off between the corresponding loss functions.
The reconstruction terms are defined as
\begin{align}
    \mathcal{L}_{\text{in}} &= \frac{1}{\norm{E_t}_1} \norm{E_t \odot (\tilde{F}_t - {F}^*_t)}_1, \\
    \mathcal{L}_{\text{out}} &= \frac{1}{\norm{1-E_t}_1} \norm{(1-E_t) \odot (\tilde{F}_t - {F}^*_t)}_1, 
\end{align}
where ${F}^*_t$ is the respective ground-truth image.
The adversarial loss follows the GAN formulation of \cite{pathak2016context}: 
\begin{equation}
    \mathcal{L}_{\text{adv}} = \max_D \; \mathbb{E}_{F_t^*}\big(\log D(F_t^*)\big) + \mathbb{E}_{\hat{F}_t}\big(\log (1 - D(\hat{F}_t))\big)
\end{equation}
Note that we compute $ \mathcal{L}_{\text{in}} $ and $ \mathcal{L}_{\text{out}} $ on the intermediate image $\tilde{F}_t$ before blending to aid generalization of the encoder/decoder networks, and $ \mathcal{L}_{\text{adv}}$ on the final result $\hat{F}_t$ to improve visual fidelity.

We do not apply an explicit cross-view consistency loss, as the upstream reconstruction stage (RIFTCast) has already aggregated the available multi-view information in the current timestep.
Instead, our model relies on reprojection and RoPE-based spatio-temporal attention to exploit cross-view context where it remains available in the rendered inputs.

During inference in the context of 3D streaming, we can make use of the fact that the video streams arrive frame by frame and that the encoder $\mathcal{E}$ receives the same frames multiple times.
By caching the encoded feature maps up to frame $t-k_w  n_w$, we significantly reduce the recomputation of values if enough memory is available.

\section{Evaluation}

\begin{table*}[]
    \centering
    \fontsize{9pt}{9pt}\selectfont
    \begin{tabular}{lc|cccc|ccc|c}
    \toprule
         & & \multicolumn{4}{c|}{Whole Image} & \multicolumn{3}{c|}{Inpainted Regions} & \\
        Model & Variant  & PSNR $\uparrow$ & SSIM $\uparrow$ & LPIPS $\downarrow$ & VFID $\downarrow$  & PSNR $\uparrow$ & SSIM $\uparrow$ & LPIPS $\downarrow$ & FPS $\uparrow$ \\
         \midrule
         \midrule

%%%%%%%%%%%%%%%%%%%%%%%%%%%%%%%%%%%%%%%%%%%%%%%%
%%%%%%%%%%%%%%%%%%%%%%%%%%%%%%%%%%%%%%%%%%%%%%%%
%%%%%%%%%%%%%%%%%%%%%%%%%%%%%%%%%%%%%%%%%%%%%%%%

\multicolumn{2}{l|}{\textbf{RGVI} \cite{cho2025elevating} \quad -- \quad  \emph{offline} \quad --} & $33.478$ & $0.9881$ & $0.0140$ & $1.6203$ & $42.582$ & $0.99834$ & $0.00678$ & $\phantom{0}3.09$ \\

\midrule
\midrule

\textbf{DSTT}  \cite{Liu_2021_DSTT} & def & $31.532$ & $0.9827$ & $0.0332$ & $3.2139$ & $35.091$ & $0.99733$ & $0.00761$ & $12.82$ \\

\textit{pretrained} & win & $31.986$ & $0.9835$ & $0.0317$ & $2.4210$ & $36.252$ & $0.99788$ & $0.00477$ & $\phantom{0}5.96$ \\

& mul & $32.045$ & $0.9835$ & $0.0316$ & $2.3864$ & $36.490$ & \colorbox{second}{$0.99793$} & $0.00470$ & $\phantom{0}1.71$ \\ 
  
\midrule
  
\textbf{DSTT} \cite{Liu_2021_DSTT} & def  & $31.341$ & $0.9825$ & $0.0341$ & $3.4281$ & $34.972$ & $0.99731$ & $0.00747$ & \colorbox{second}{$13.68$} \\

\textit{finetuned} & win & $31.913$ & $0.9833$ & $0.0317$ & $2.8144$ & \colorbox{second}{$37.642$} & $0.99770$ & $0.00889$ & $\phantom{0}3.40$ \\
  
& mul & $31.937$ & $0.9833$ & $0.0326$ & $2.5660$ & $36.391$ & $0.99789$ & $0.00473$ & $\phantom{0}1.70$ \\ 

\midrule

\textbf{Fuseformer}  \cite{Liu_2021_FuseFormer} & def & $31.884$ & $0.9832$ & $0.0316$ & $3.1830$ & $35.421$ & $0.99749$ & $0.00646$ & $\phantom{0}8.64$ \\

\textit{pretrained} & win & $32.050$ & $0.9836$ & $0.0310$ & $2.2516$ & $36.107$ & $0.99785$ & $0.00473$ & $\phantom{0}3.61$ \\

& mul & \colorbox{second}{$32.156$} & \colorbox{second}{$0.9838$} & \colorbox{second}{$0.0303$} & $2.2095$ & $36.371$ & $0.99791$ & \colorbox{second}{$0.00458$} & $\phantom{0}0.76$ \\ 

\midrule

\textbf{Fuseformer}  \cite{Liu_2021_FuseFormer} & def & $31.726$ & $0.9829$ & $0.0325$ & $3.3018$ & $35.317$ & $0.99746$ & $0.00672$ & $\phantom{0}7.50$ \\

 \textit{finetuned}  & win & $31.930$ & $0.9834$ & $0.0323$ & $2.5002$ & $36.090$ & $0.99785$ & $0.00477$ & $\phantom{0}3.37$ \\
 
  & mul & $31.978$ & $0.9835$ & $0.0320$ & $2.4637$ & $36.236$ & $0.99789$ & $0.00473$ & $\phantom{0}0.75$ \\
  
  \midrule
  
\textbf{E2FGVI}  \cite{liCvpr22vInpainting} & def & $31.834$ & $0.9831$ & $0.0320$ & $3.1350$ & $35.444$ & $0.99750$ & $0.00666$ & $\phantom{0}6.14$ \\

 \textit{pretrained}  & win & $32.055$ & $0.9837$ & $0.0311$ & $2.2489$ & $36.210$ & $0.99788$ & $0.00467$ & $\phantom{0}3.07$ \\
 
& mul & $32.155$ & \colorbox{second}{$0.9838$} & $0.0306$ & \colorbox{second}{$2.1975$} & $36.430$ & $0.99791$ & \colorbox{second}{$0.00458$} & $\phantom{0}0.90$ \\
  
\midrule
  
\textbf{E2FGVI}  \cite{liCvpr22vInpainting} & def  & $31.908$ & $0.9832$ & $0.0314$ & $3.1274$ & $35.299$ & $0.99744$ & $0.00690$ & $\phantom{0}7.10$ \\

\textit{finetuned} & win  & $31.895$ & $0.9833$ & $0.0324$ & $2.5528$ & $35.769$ & $0.99771$ & $0.00527$ & $\phantom{0}3.07$ \\

& mul & $31.879$ & $0.9832$ & $0.0325$ & $2.5562$ & $35.742$ & $0.99766$ & $0.00541$ & $\phantom{0}0.92$ \\

\midrule

\textbf{Ours} &  & \colorbox{best}{$32.616$} & \colorbox{best}{$0.9851$} & \colorbox{best}{$0.0262$} & \colorbox{best}{$1.6671$} & \colorbox{best}{$42.184$} & \colorbox{best}{$0.99911$} & \colorbox{best}{$0.00224$} & \colorbox{best}{$41.55$} \\
    \bottomrule
    \end{tabular}
    \caption{Results of all baseline methods compared to our method on various metrics, measured either on the whole image or only for pixels belonging to the inpainted regions. DSTT, FuseFormer, and E2FGVI are reported in three variants: Default settings of the pretrained model (def), windowed approach (win) and multiple views as an input (mul). Colored boxes show \colorbox{best}{best} and \colorbox{second}{second-best} results per metric of online methods. RGVI is included for reference.}
    \label{tab:results}
\end{table*}

\begin{table*}[]
    \centering
    \fontsize{9pt}{9pt}\selectfont
    \begin{tabular}{l|cccc|ccc|c}
    \toprule
         & \multicolumn{4}{c|}{Whole Image} & \multicolumn{3}{c|}{Inpainted Regions} & \\
        Model \hspace{35mm} & PSNR $\uparrow$ & SSIM $\uparrow$ & LPIPS $\downarrow$ & VFID $\downarrow$  & PSNR $\uparrow$ & SSIM $\uparrow$ & LPIPS $\downarrow$ & FPS $\uparrow$ \\
         \midrule
         \midrule

\textbf{DSTT}  \cite{Liu_2021_DSTT} & \colorbox{second}{$32.977$} & $0.9822$ & $0.0159$ & $1.4968$ & $36.051$ & $0.9967$ & $0.0109$ & $\phantom{0}0.71$ \\
\textbf{Fuseformer}  \cite{Liu_2021_FuseFormer} & $32.678$ & \colorbox{second}{$0.9832$} & \colorbox{second}{$0.0206$} & $1.4056$ & $37.361$ & $0.9975$ & $0.0095$ & $\phantom{0}0.69$ \\ 
\textbf{E2FGVI}  \cite{liCvpr22vInpainting} & $32.766$ & \colorbox{best}{$0.9833$} & \colorbox{best}{$0.0204$} & \colorbox{second}{$1.3748$} & \colorbox{second}{$37.535$} & \colorbox{second}{$0.9976$} & \colorbox{second}{$0.0092$} & \colorbox{second}{$\phantom{0}0.82$} \\

% \midrule

\midrule
\textbf{Ours} & \colorbox{best}{$33.059$} & $0.9800$ & $0.0287$ & \colorbox{best}{$0.9953$} & \colorbox{best}{$42.192$} & \colorbox{best}{$0.9989$} & \colorbox{best}{$0.0031$} & \colorbox{best}{$37.01$} \\
    \bottomrule
    \end{tabular}
    \caption{Results of the online baseline methods (pretrained, mul) compared to our method (without fine-tuning) on various metrics on the RIFTCast dataset~\cite{zingsheim2025riftcast}. Colored boxes show \colorbox{best}{best} and \colorbox{second}{second-best} results per metric.}
    \label{tab:results_bonn}
\end{table*}

\begin{figure*}[t]
    \newcommand{\percellwidth}{0.135\linewidth}
    \centering
    \captionsetup[subfigure]{justification=centering}
    
    \def\zoomx{-0.65}
    \def\zoomy{0.70}
    \def\zoomboxx{0.6}
    \def\zoomboxy{-1.25}
    \def\zoomlevel{1.6}

    \begin{subfigure}[t]{\percellwidth}
        \imagewithzoom{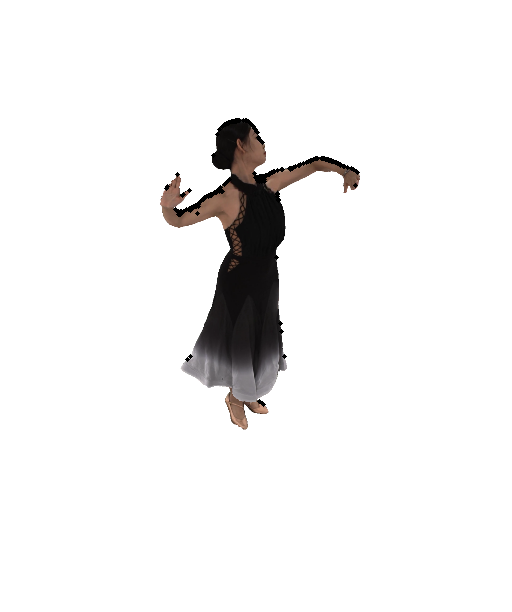}{\zoomx}{\zoomy}{\zoomboxx}{\zoomboxy}{1.2cm}{\zoomlevel}{150 180 150 100}
    \end{subfigure}
    \hfill
    \begin{subfigure}[t]{\percellwidth}
        \imagewithzoom{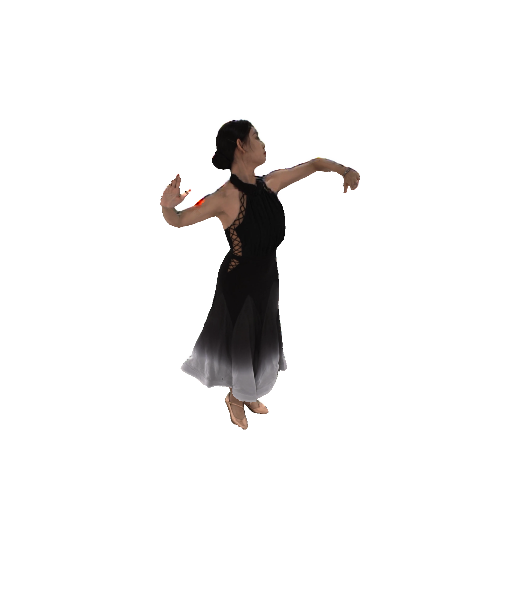}{\zoomx}{\zoomy}{\zoomboxx}{\zoomboxy}{1.2cm}{\zoomlevel}{150 180 150 100}
    \end{subfigure}
    \hfill
    \begin{subfigure}[t]{\percellwidth}
        \imagewithzoom{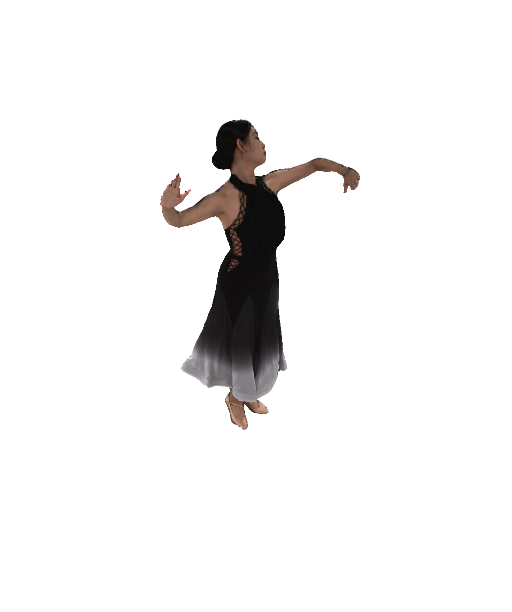}{\zoomx}{\zoomy}{\zoomboxx}{\zoomboxy}{1.2cm}{\zoomlevel}{150 180 150 100}
    \end{subfigure}
    \hfill
    \begin{subfigure}[t]{\percellwidth}
        \imagewithzoom{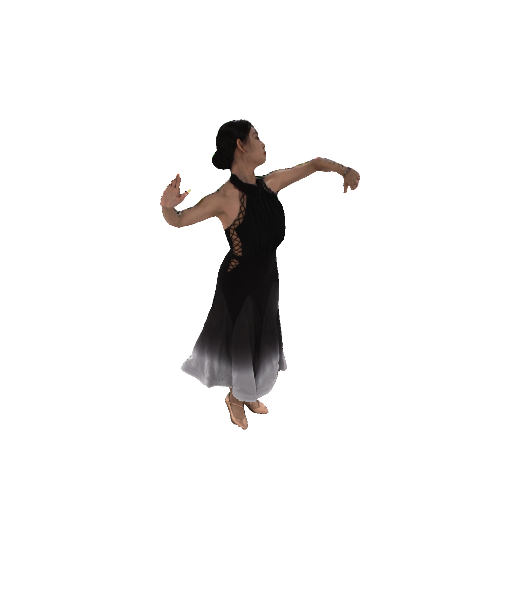}{\zoomx}{\zoomy}{\zoomboxx}{\zoomboxy}{1.2cm}{\zoomlevel}{150 180 150 100}
    \end{subfigure}
    \hfill
    \begin{subfigure}[t]{\percellwidth}
        \imagewithzoom{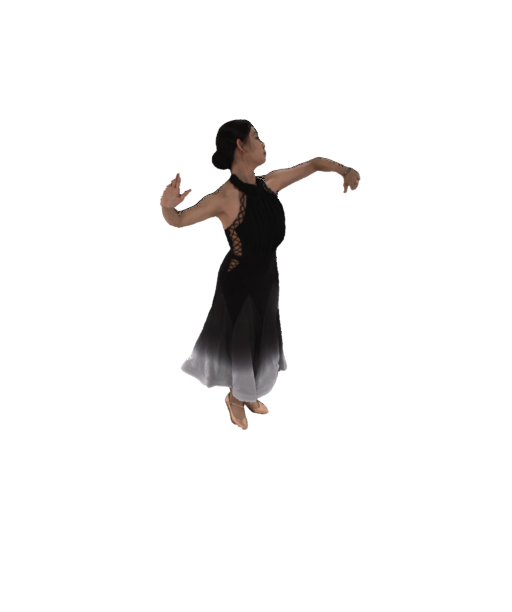}{\zoomx}{\zoomy}{\zoomboxx}{\zoomboxy}{1.2cm}{\zoomlevel}{150 180 150 100}
    \end{subfigure}
    \hfill
    \begin{subfigure}[t]{\percellwidth}
        \imagewithzoom{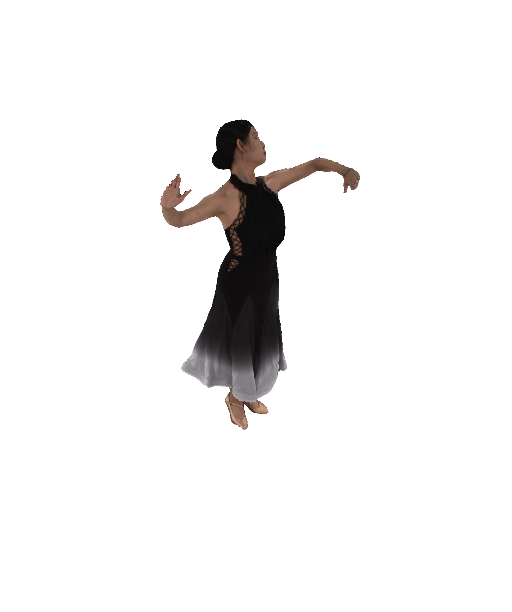}{\zoomx}{\zoomy}{\zoomboxx}{\zoomboxy}{1.2cm}{\zoomlevel}{150 180 150 100}
    \end{subfigure}
    \hfill
    \begin{subfigure}[t]{\percellwidth}
        \imagewithzoom{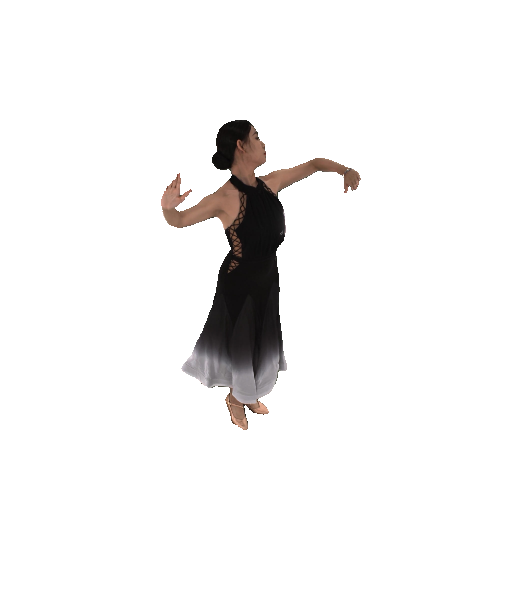}{\zoomx}{\zoomy}{\zoomboxx}{\zoomboxy}{1.2cm}{\zoomlevel}{150 180 150 100}
    \end{subfigure}

    % %%%%%%%%%%%%%%%%%%%%%%%%%

    \def\zoomx{-0.65}
    \def\zoomy{0.82}
    \def\zoomboxx{-0.6}
    \def\zoomboxy{-1.05}
    \def\zoomlevel{2}

    \begin{subfigure}[t]{\percellwidth}
        \imagewithzoom{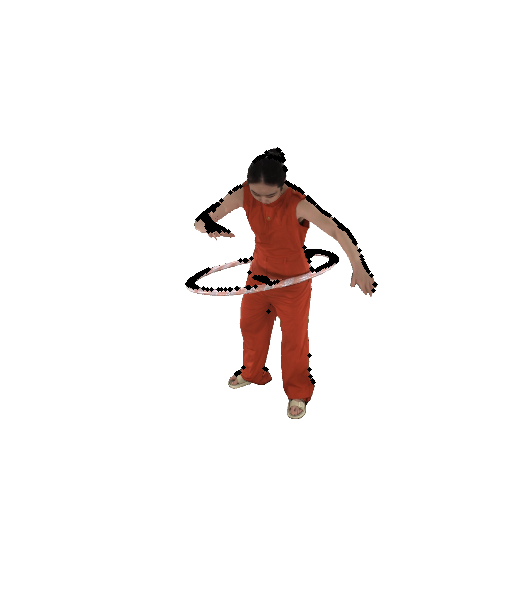}{\zoomx}{\zoomy}{\zoomboxx}{\zoomboxy}{1.2cm}{\zoomlevel}{170 180 130 140}
    \end{subfigure}
    \hfill
    \begin{subfigure}[t]{\percellwidth}
        \imagewithzoom{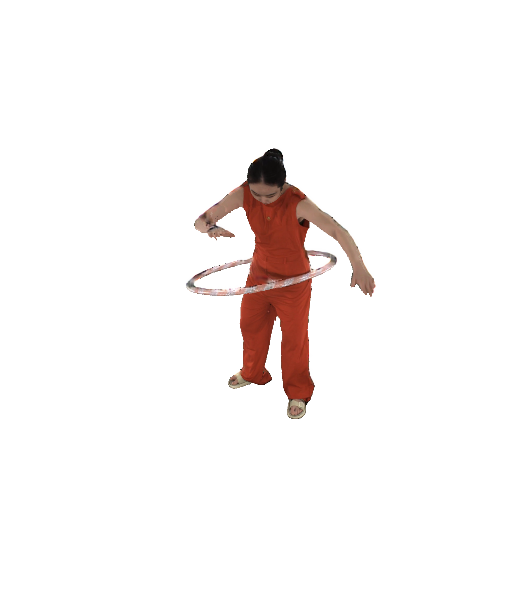}{\zoomx}{\zoomy}{\zoomboxx}{\zoomboxy}{1.2cm}{\zoomlevel}{170 180 130 140}
    \end{subfigure}
    \hfill
    \begin{subfigure}[t]{\percellwidth}
        \imagewithzoom{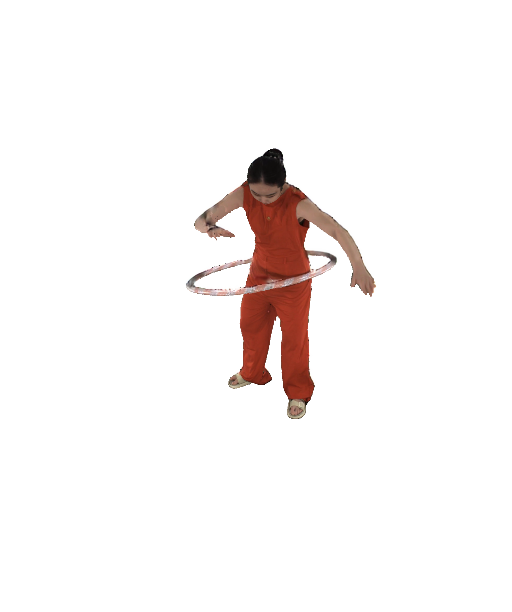}{\zoomx}{\zoomy}{\zoomboxx}{\zoomboxy}{1.2cm}{\zoomlevel}{170 180 130 140}
    \end{subfigure}
    \hfill
    \begin{subfigure}[t]{\percellwidth}
        \imagewithzoom{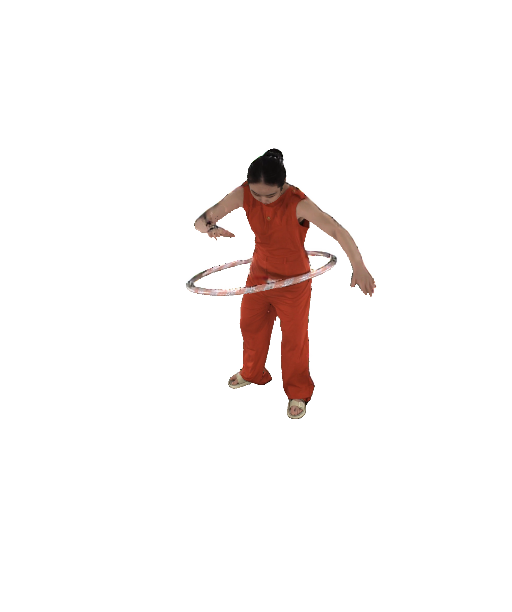}{\zoomx}{\zoomy}{\zoomboxx}{\zoomboxy}{1.2cm}{\zoomlevel}{170 180 130 140}
    \end{subfigure}
    \hfill
    \begin{subfigure}[t]{\percellwidth}
        \imagewithzoom{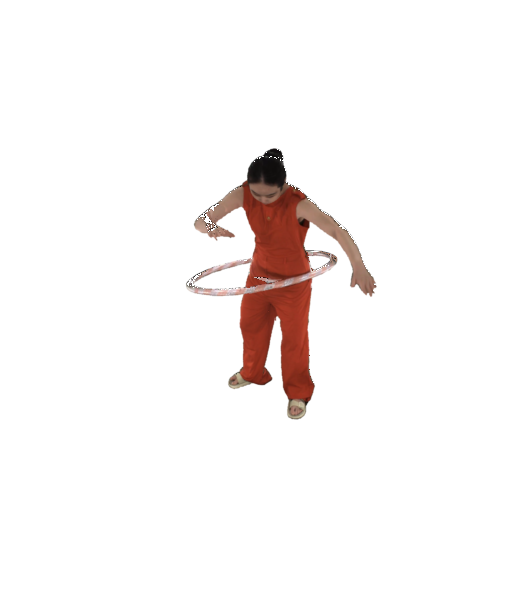}{\zoomx}{\zoomy}{\zoomboxx}{\zoomboxy}{1.2cm}{\zoomlevel}{170 180 130 140}
    \end{subfigure}
    \hfill
    \begin{subfigure}[t]{\percellwidth}
        \imagewithzoom{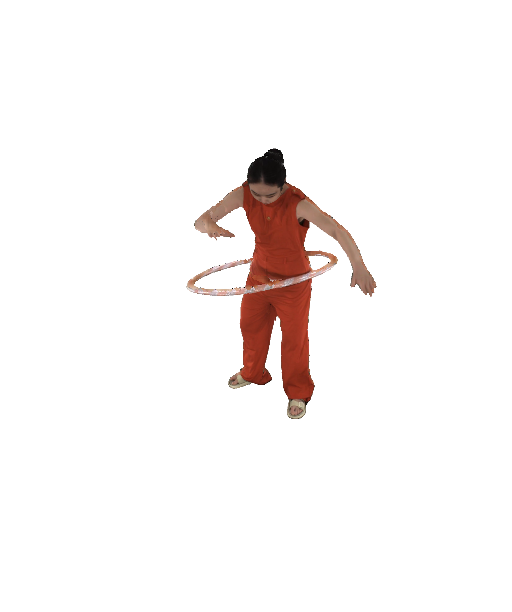}{\zoomx}{\zoomy}{\zoomboxx}{\zoomboxy}{1.2cm}{\zoomlevel}{170 180 130 140}
    \end{subfigure}
    \hfill
    \begin{subfigure}[t]{\percellwidth}
        \imagewithzoom{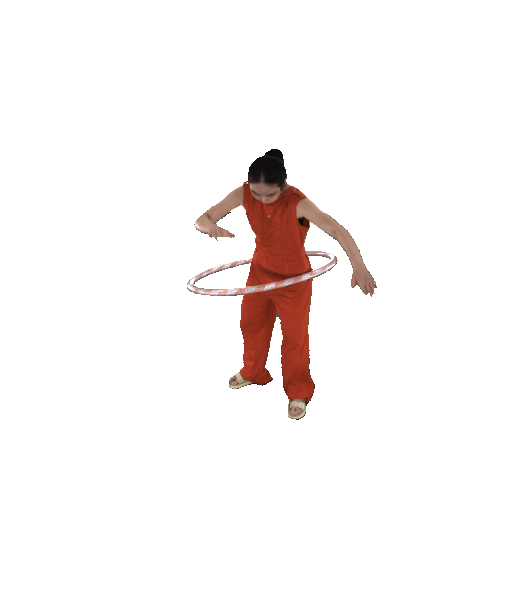}{\zoomx}{\zoomy}{\zoomboxx}{\zoomboxy}{1.2cm}{\zoomlevel}{170 180 130 140}
    \end{subfigure}

    % %%%%%%%%%%%%%%%%%%%%%%%%%

    \def\zoomx{-0.1}
    \def\zoomy{-1.37}
    \def\zoomboxx{0.58}
    \def\zoomboxy{-0.2}
    \def\zoomlevel{2.5}

    \begin{subfigure}[t]{\percellwidth}
        \imagewithzoom{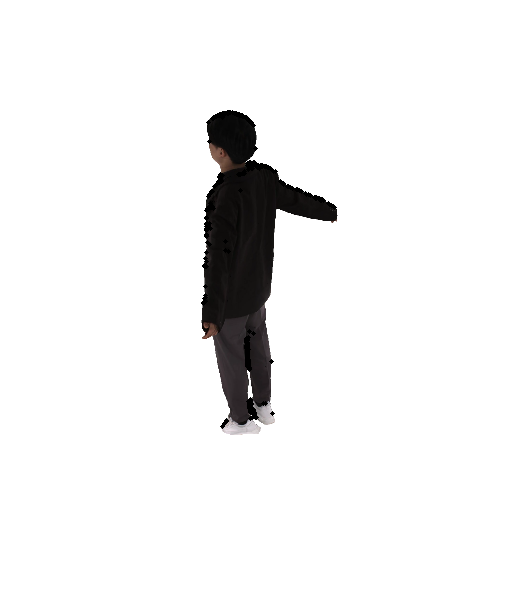}{\zoomx}{\zoomy}{\zoomboxx}{\zoomboxy}{1.2cm}{\zoomlevel}{140 160 140 100}
        \caption*{Rendered Image}
    \end{subfigure}
    \hfill
    \begin{subfigure}[t]{\percellwidth}
        \imagewithzoom{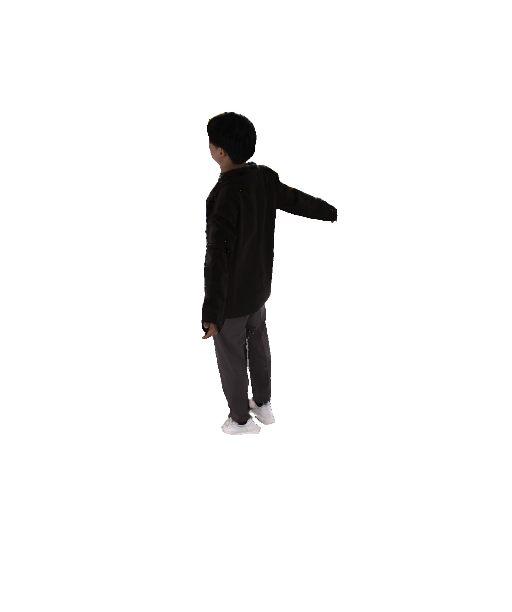}{\zoomx}{\zoomy}{\zoomboxx}{\zoomboxy}{1.2cm}{\zoomlevel}{140 160 140 100}
        \caption*{DSTT \cite{Liu_2021_DSTT}}
    \end{subfigure}
    \hfill
    \begin{subfigure}[t]{\percellwidth}
        \imagewithzoom{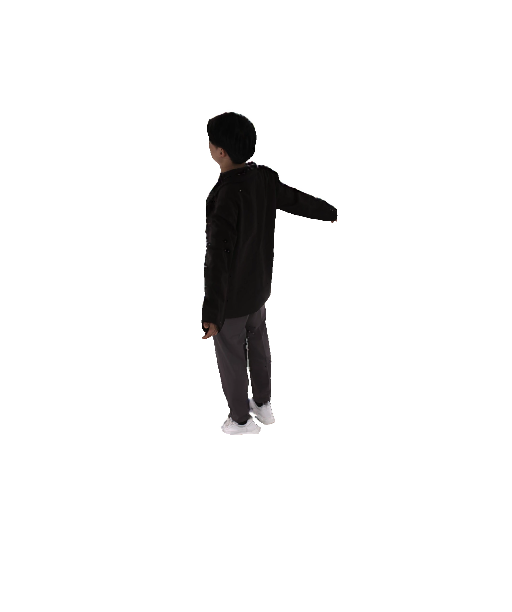}{\zoomx}{\zoomy}{\zoomboxx}{\zoomboxy}{1.2cm}{\zoomlevel}{140 160 140 100}
        \caption*{FuseFormer \cite{Liu_2021_FuseFormer}}
    \end{subfigure}
    \hfill
    \begin{subfigure}[t]{\percellwidth}
        \imagewithzoom{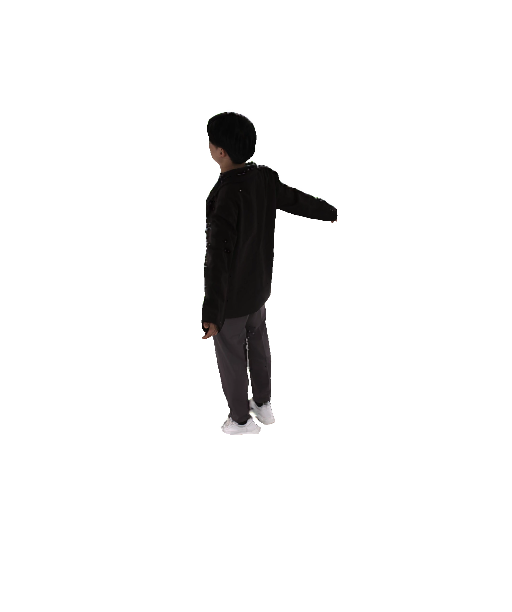}{\zoomx}{\zoomy}{\zoomboxx}{\zoomboxy}{1.2cm}{\zoomlevel}{140 160 140 100}
        \caption*{E2FGVI \cite{liCvpr22vInpainting}}
    \end{subfigure}
    \hfill
    \begin{subfigure}[t]{\percellwidth}
        \imagewithzoom{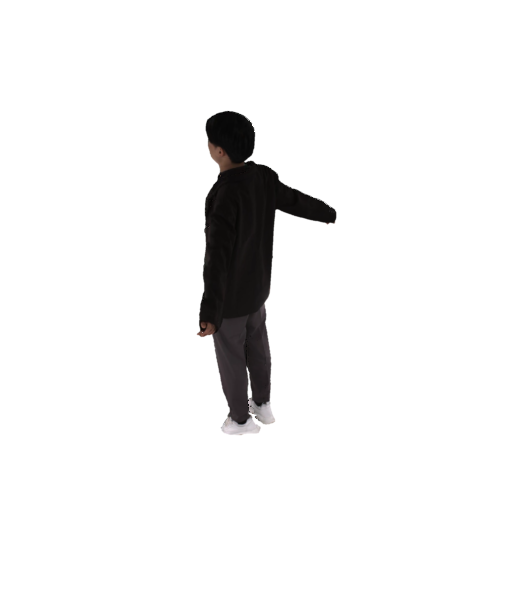}{\zoomx}{\zoomy}{\zoomboxx}{\zoomboxy}{1.2cm}{\zoomlevel}{140 160 140 100}
        \caption*{RGVI \cite{cho2025elevating}}
    \end{subfigure}
    \hfill
    \begin{subfigure}[t]{\percellwidth}
        \imagewithzoom{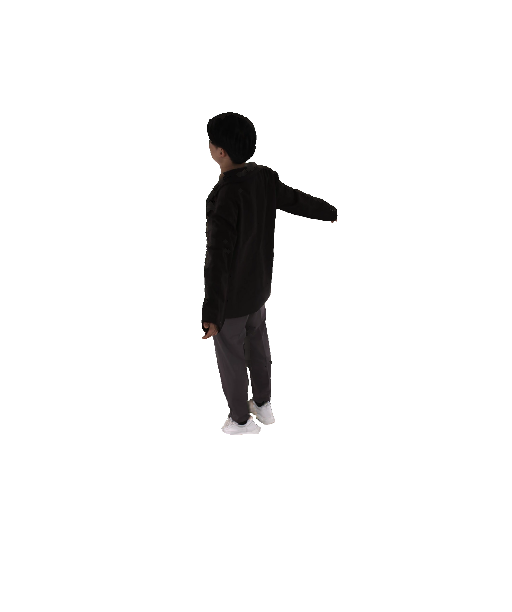}{\zoomx}{\zoomy}{\zoomboxx}{\zoomboxy}{1.2cm}{\zoomlevel}{140 160 140 100}
        \caption*{Ours}
    \end{subfigure}
    \hfill
    \begin{subfigure}[t]{\percellwidth}
        \imagewithzoom{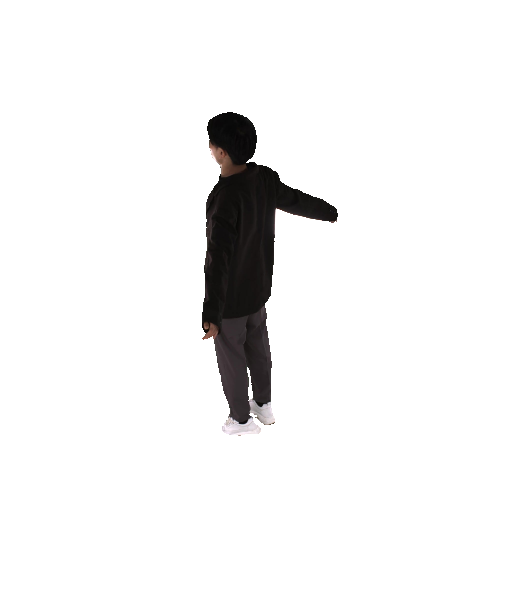}{\zoomx}{\zoomy}{\zoomboxx}{\zoomboxy}{1.2cm}{\zoomlevel}{140 160 140 100}
        \caption*{Ground Truth}
    \end{subfigure}

    \caption{Visual comparison of our method against the pretrained multicam variants of the baseline methods. First column shows the input image from the reconstruction framework, last column shows the ground-truth view seen from the omitted camera.}
    \label{fig:results}
\end{figure*}

We trained and evaluated our method on a subset of the DNARendering dataset~\cite{2023dnarendering}, a real-world dataset of dynamic human performances, and randomly sampled $72$ scenes for training and $7$ scenes for evaluation, ensuring that no subject appears in both sets.
As a reference 3D streaming method, we use RIFTCast~\cite{zingsheim2025riftcast} which utilizes foreground masks to build a visual hull as the geometry proxy $\mathcal{G}_t$. 
Then, given a novel view camera, a subset of input views that is close to the target view is selected to comply with real-time constraints.
In practice, we used exactly the same three closest input views as in RIFTCast to synthesize the target novel view $F_t$ and provide them (together with past frames) as input to our inpainting model.

To show the generalization ability of our method, we additionally tested our method on their multi-view dataset, consisting of 31 scenes captured with 34 synchronized RGB cameras featuring complex dynamic multi-actor and actor-object interactions.

For training and evaluation, we excluded one randomly selected camera from the inputs of the 3D streaming and used it as ground truth for comparison. 
The quality is measured using common image and video metrics: peak signal-to-noise ratio (PSNR), structural similarity index (SSIM)~\cite{wang2004image}, and learned perceptual image patch similarity (LPIPS)~\cite{zhang2018perceptual}, as well as the video Fréchet inception distance (VFID)~\cite{wang2018video}.
We evaluated PSNR, SSIM, and LPIPS on the whole image, which also includes errors from the streaming method.
In addition, we also report these values for only inpainted pixels, where SSIM and LPIPS are computed between masked versions of the result and ground-truth images, whereas PSNR is averaged over only the inpainted pixels.

\subsection{Methods and Implementation}
In our experiments, we set the parameters that define the number of additional timesteps auxiliary to the current frame to ${n_w=n_c=3}$ and ${k_w=10}$, which makes the context contain at most 7 frames per camera.
The patch size is set to be ${7\times 7}$, with 3 pixels overlap. 
The weights for the image and adversarial losses are set to ${\lambda_{\mathrm{img}} = 1.0}$ and ${\lambda_{\mathrm{adv}} = 0.01}$.
Finally, we used ${n_{g} = 2}$ transformer groups, each composed of ${n_{b} = 4}$ blocks, in our pipeline.

We compare our method against several variants of three recent, efficient inpainting models~\cite{Liu_2021_DSTT,Liu_2021_FuseFormer,liCvpr22vInpainting} that were adapted to be used as online approaches by Thiry \etal~\cite{thiry2024towards}.
For a fair comparison, we performed the evaluation on different variants of the baseline methods:
\begin{itemize}
    \item \textbf{Default (def):} This resembles the unmodified baseline where images are processed in their lower, architecture-specific resolution and then upscaled.
    \item \textbf{Windowed (win):} We modified the inference step by first splitting the frames into individual overlapping windows matching the native resolution and subsequently stitch together the predicted results.
    \item \textbf{Multi-View (mul):} For a fair comparison, we adapted the baselines to also receive frames from multiple cameras, similar to our approach.
    Since their architectures are single-view by design, we interleaved additional camera views into the temporal input sequence.
    Although this is not an ideal use of multi-view information, it ensures that all methods are evaluated with comparable input data.
\end{itemize}
In addition, we also evaluated versions of the models that are finetuned on the same DNARendering subset used to train our model.
The training follows the parameters and methodologies provided by the respective authors, resulting in six variants per baseline method.
To understand the performance gap to offline methods, we also added results from RGVI \cite{cho2025elevating}.
As this method does not support multi-view inputs, we instead inferenced it with ground-truth frames from the target view up to the current inpainting frame.

\subsection{Quantitative and Qualitative Results}

\begin{table*}[]
    \centering
    \fontsize{9pt}{9pt}\selectfont
    \begin{tabular}{l|c|c|c|c|c|c|c|c}
       \toprule
       & \multicolumn{4}{c|}{Whole Image} & \multicolumn{3}{c|}{Inpainted Regions} & \\
        & PSNR $\uparrow$ & SSIM $\uparrow$ & LPIPS $\downarrow$ & VFID $\downarrow$  & PSNR $\uparrow$ & SSIM $\uparrow$ & LPIPS $\downarrow$ & FPS $\uparrow$ \\
        \midrule
        \midrule
        Single cam w/o masks & \colorbox{second}{$32.400$} & \colorbox{second}{$0.9846$} & \colorbox{best}{$0.0261$} & \colorbox{best}{$1.6009$} & $34.426$ & $0.9971$ & $0.0092$ & \colorbox{best}{$87.46$} \\
        Single cam & $32.277$ & $0.9843$ & $0.0275$ & $1.7331$ & \colorbox{second}{$40.004$} & \colorbox{second}{$0.9986$} & \colorbox{second}{$0.0034$} & \colorbox{second}{$79.43$} \\
        w/o temp & $32.237$ & $0.9843$ & $0.0275$ & $1.7320$ & $39.838$ & \colorbox{second}{$0.9986$} & $0.0039$ & $41.32$ \\
        w/o RoPE & $32.046$ & $0.9839$ & $0.0278$ & $1.7963$ & $38.920$ & $0.9983$ & $0.0050$ & $41.05$ \\
        \midrule
        Ours & \colorbox{best}{$32.616$} & \colorbox{best}{$0.9851$} & \colorbox{second}{$0.0262$} & \colorbox{second}{$1.6671$} & \colorbox{best}{$42.184$} & \colorbox{best}{$0.9991$} & \colorbox{best}{$0.0022$} & $41.55$ \\
        % no context 
       \bottomrule
    \end{tabular}
    \caption{Ablation study on the impact of key components of our method. Colored boxes show \colorbox{best}{best} and \colorbox{second}{second} results per metric.}
    \label{tab:ablation}
\end{table*}

\begin{figure*}[t]
    \newcommand{\percellwidth}{0.14\linewidth}
    \centering
    \captionsetup[subfigure]{justification=centering}
    
    \def\zoomx{0.75}
    \def\zoomy{0.98}
    \def\zoomboxx{-0.63}
    \def\zoomboxy{-0.93}
    \def\zoomlevel{1.6}

    \begin{subfigure}[t]{\percellwidth}
        \imagewithzoom{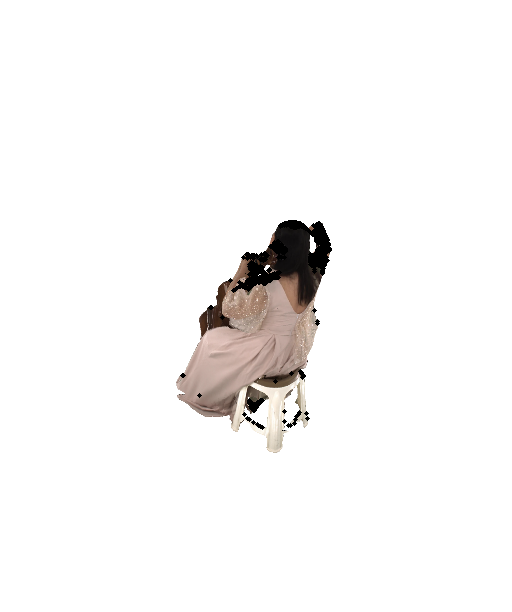}{\zoomx}{\zoomy}{\zoomboxx}{\zoomboxy}{1.2cm}{\zoomlevel}{170 190 170 210}
        \caption*{Input}
    \end{subfigure}
    \hfill
    \begin{subfigure}[t]{\percellwidth}
        \imagewithzoom{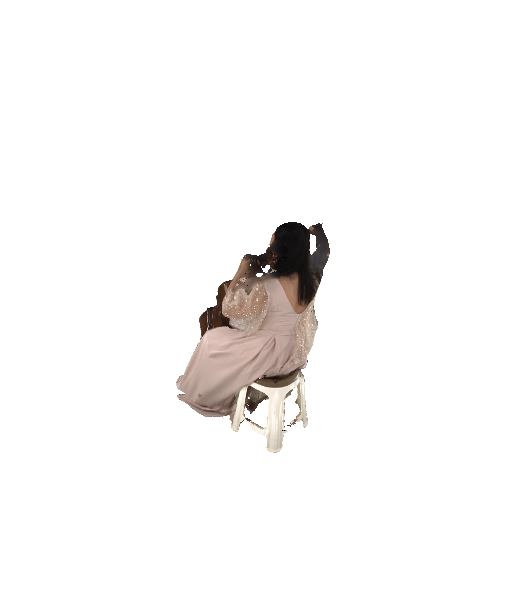}{\zoomx}{\zoomy}{\zoomboxx}{\zoomboxy}{1.2cm}{\zoomlevel}{170 190 170 210}
        \caption*{Single cam w/o masks}
    \end{subfigure}
    \hfill
    \begin{subfigure}[t]{\percellwidth}
        \imagewithzoom{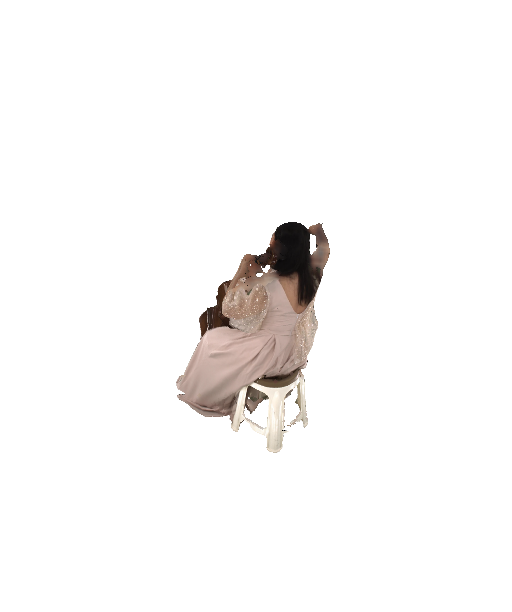}{\zoomx}{\zoomy}{\zoomboxx}{\zoomboxy}{1.2cm}{\zoomlevel}{170 190 170 210}
        \caption*{Single cam}
    \end{subfigure}
    \hfill
    \begin{subfigure}[t]{\percellwidth}
        \imagewithzoom{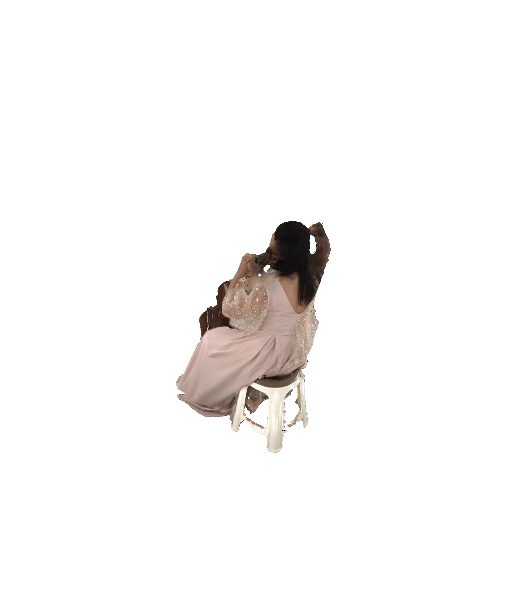}{\zoomx}{\zoomy}{\zoomboxx}{\zoomboxy}{1.2cm}{\zoomlevel}{170 190 170 210}
        \caption*{w/o temp}
    \end{subfigure}
    \hfill
    \begin{subfigure}[t]{\percellwidth}
        \imagewithzoom{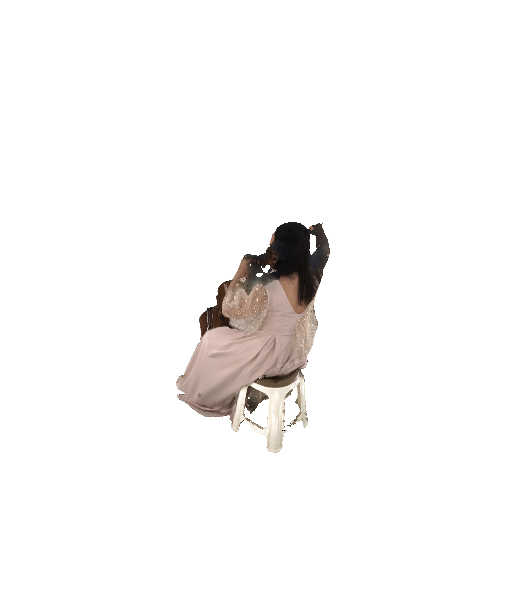}{\zoomx}{\zoomy}{\zoomboxx}{\zoomboxy}{1.2cm}{\zoomlevel}{170 190 170 210}
        \caption*{w/o RoPE}
    \end{subfigure}
    \hfill
    \begin{subfigure}[t]{\percellwidth}
        \imagewithzoom{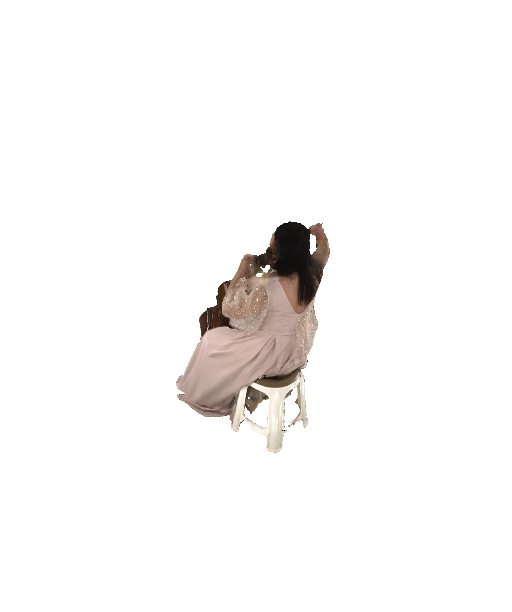}{\zoomx}{\zoomy}{\zoomboxx}{\zoomboxy}{1.2cm}{\zoomlevel}{170 190 170 210}
        \caption*{Ours}
    \end{subfigure}

    \caption{Ablation study showing  (from left to right): the input image, reconstruction using only a single camera view once without and once with masks, without leveraging past video frames, without Rotary Positional Encodings (RoPE), and the full proposed pipeline.}
    \label{fig:ablation}
\end{figure*}

\Cref{tab:results} presents the key findings of our evaluation. 
While the windowed (win) and multi-view (mul) approaches increase the perceptual quality, especially in the inpainted regions, they come at the cost of slower inference.
Fine-tuning on the DNARendering dataset did not yield improvements and, in most metrics, resulted in a slight decline in quality.
This might be due to keeping the original training setup for each baseline, which is not necessarily tailored for images without background or thin inpainting regions along object boundaries.
Also, RGVI sometimes fails to inpaint with foreground pixels and falls back to filling with the white background (Fig. 4, row 2).
In contrast, our method is able to significantly outperform the baseline methods across all metrics.
Note that PSNR values are generally higher when restricted to inpainted regions compared to the whole image.
This is because the full-image score also reflects reconstruction errors from the streaming, $F_t$, while the evaluation masked with $E_t$ isolates only the regions directly optimized by our model.
We also show a qualitative comparison between our method and the baselines in \Cref{fig:results}.
Here, the baselines often inpaint larger regions with dark colors or introduce color artifacts.
This is the case for the arm in the first sample, where a glowing red dot is visible in the DSTT result.
Likewise, in the second sample all baseline methods produce gray artifacts across the arm, while our method reproduces the skin color more faithfully.
These artifacts are also visible in the third example, where the tip of the right shoe is occluded.
Furthermore, baseline methods blur the dark color of the pants with the white shoe, whereas our model generates a clearer boundary.

\subsection{Generalization Capabilities}

\begin{figure}[t]

    \newcommand{\percellwidth}{0.3\linewidth}
    \centering
    \captionsetup[subfigure]{justification=centering}

    \def\zoomx{0.45}
    \def\zoomy{-0.6}
    \def\zoomboxx{-0.65}
    \def\zoomboxy{0.55}
    \def\zoomlevel{3}

    \begin{subfigure}[t]{\percellwidth}
        \imagewithzoom{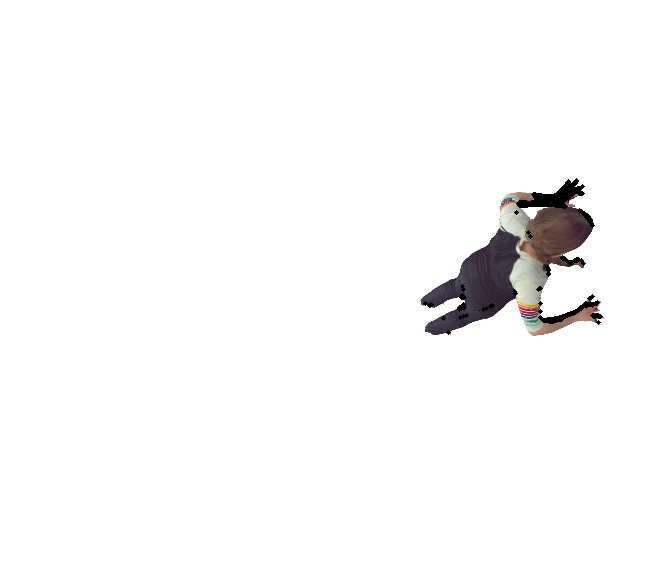}{\zoomx}{\zoomy}{\zoomboxx}{\zoomboxy}{1.2cm}{\zoomlevel}{410 220 50 170}
    \end{subfigure}
    \hfill
    \begin{subfigure}[t]{\percellwidth}
        \imagewithzoom{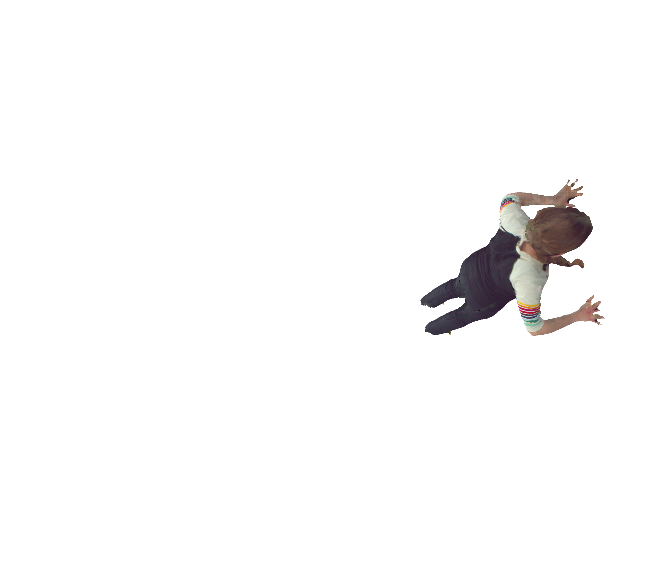}{\zoomx}{\zoomy}{\zoomboxx}{\zoomboxy}{1.2cm}{\zoomlevel}{410 220 50 170}
    \end{subfigure}
    \hfill
    \begin{subfigure}[t]{\percellwidth}
        \imagewithzoom{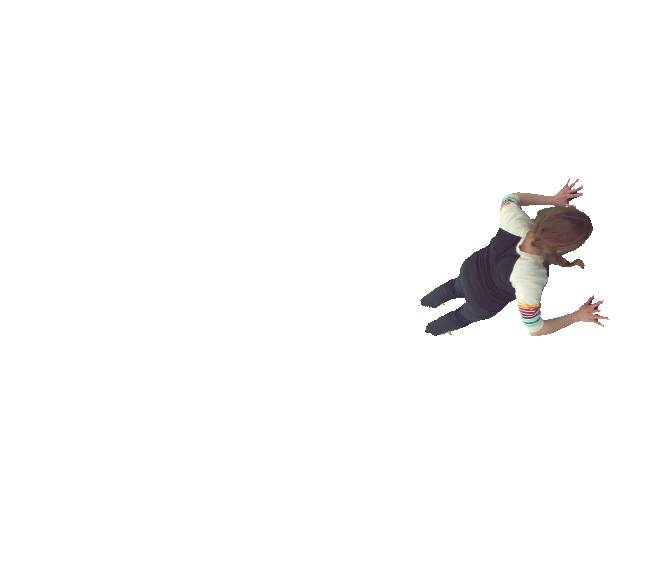}{\zoomx}{\zoomy}{\zoomboxx}{\zoomboxy}{1.2cm}{\zoomlevel}{410 220 50 170}
    \end{subfigure}
    
    \def\zoomx{-0.6}
    \def\zoomy{-0.9}
    \def\zoomboxx{-0.66}
    \def\zoomboxy{0.7}
    \def\zoomlevel{2}

    \begin{subfigure}[t]{\percellwidth}
        \imagewithzoom{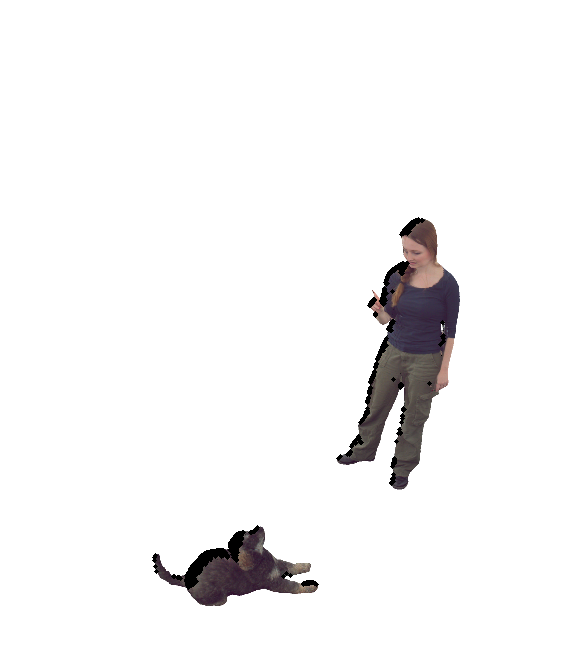}{\zoomx}{\zoomy}{\zoomboxx}{\zoomboxy}{1.2cm}{\zoomlevel}{90 20 30 150}
        \caption*{Rendered Image}
    \end{subfigure}
    \hfill
    \begin{subfigure}[t]{\percellwidth}
        \imagewithzoom{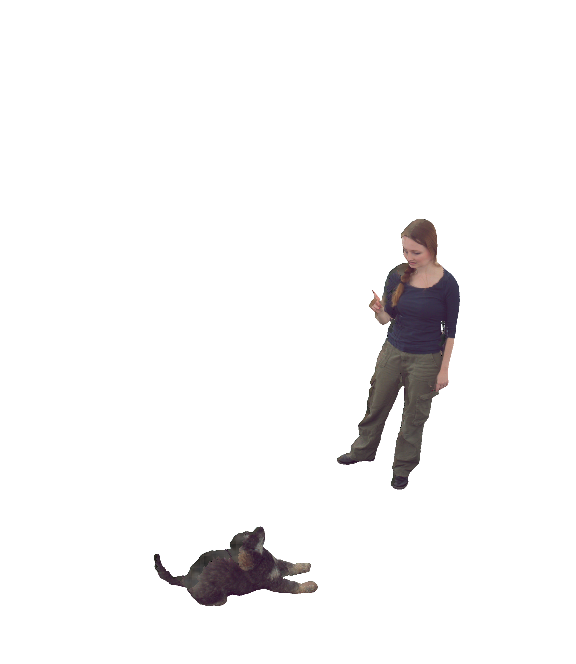}{\zoomx}{\zoomy}{\zoomboxx}{\zoomboxy}{1.2cm}{\zoomlevel}{90 20 30 150}
        \caption*{Ours}
    \end{subfigure}
    \hfill
    \begin{subfigure}[t]{\percellwidth}
        \imagewithzoom{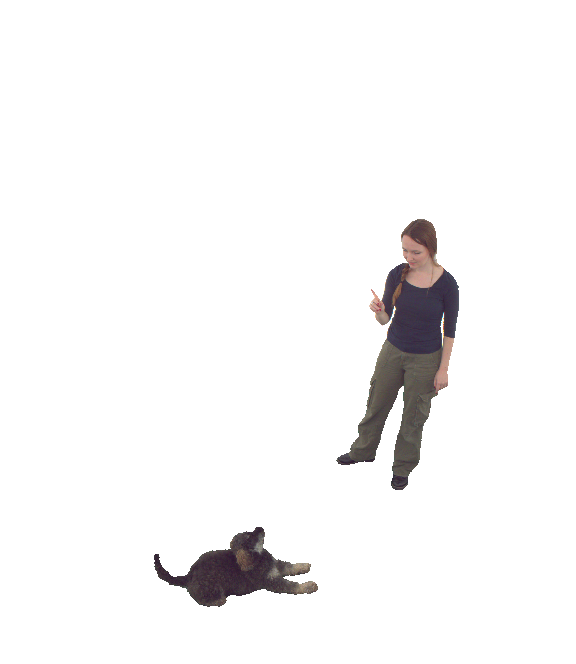}{\zoomx}{\zoomy}{\zoomboxx}{\zoomboxy}{1.2cm}{\zoomlevel}{90 20 30 150}
        \caption*{Ground Truth}
    \end{subfigure}

    \caption{Results on the RIFTCast dataset \cite{zingsheim2025riftcast}.}
    % \vspace{-12pt}
    \label{fig:qualitative_bonn}
\end{figure}

To analyze the generalization capabilities of our model, we tested it on the challenging RIFTCast dataset \cite{zingsheim2025riftcast}.
Compared to DNARendering, RIFTCast presents more challenging conditions.
Scenes frequently involve multiple actors, human–object interactions, and animals, with more complex occlusions.
Moreover, subjects move extensively within the capture volume.
These factors together make RIFTCast a substantially harder benchmark for inpainting within a streaming pipeline.
As shown in \Cref{tab:results_bonn}, without any retraining or fine-tuning, our model is able to outperform the baseline methods in the inpainted regions, similar to the comparison on the DNARendering dataset.
Note that the reduction in inference speed is due to the higher resolution of the images in the RIFTCast dataset.
Additionally, qualitative results on two scenes are shown in \Cref{fig:qualitative_bonn}.

\begin{figure}[t]

    \newcommand{\percellwidth}{0.3\linewidth}
    \centering
    \captionsetup[subfigure]{justification=centering}

    \def\zoomx{-0.2}
    \def\zoomy{0.90}
    \def\zoomboxx{0.66}
    \def\zoomboxy{-0.82}
    \def\zoomlevel{2}

    \begin{subfigure}[t]{\percellwidth}
        \imagewithzoom{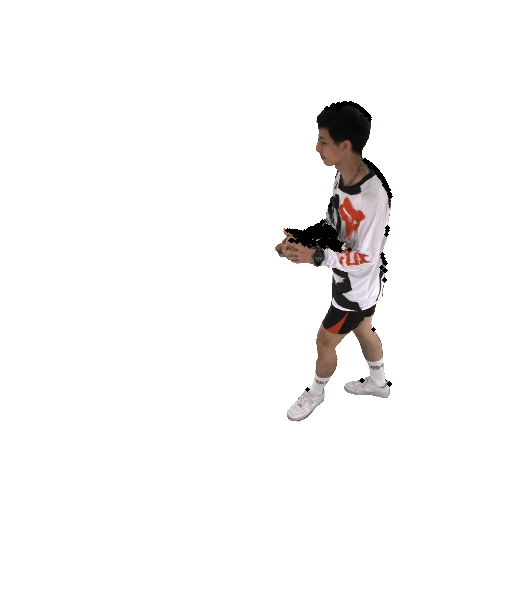}{\zoomx}{\zoomy}{\zoomboxx}{\zoomboxy}{1.2cm}{\zoomlevel}{240 190 80 200}
        \caption*{Rendered Image}
    \end{subfigure}
    \hfill
    \begin{subfigure}[t]{\percellwidth}
        \imagewithzoom{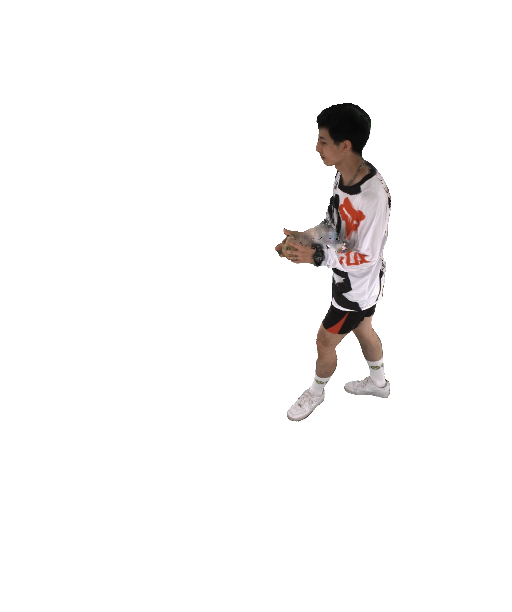}{\zoomx}{\zoomy}{\zoomboxx}{\zoomboxy}{1.2cm}{\zoomlevel}{240 190 80 200}
        \caption*{Ours}
    \end{subfigure}
    \hfill
    \begin{subfigure}[t]{\percellwidth}
        \imagewithzoom{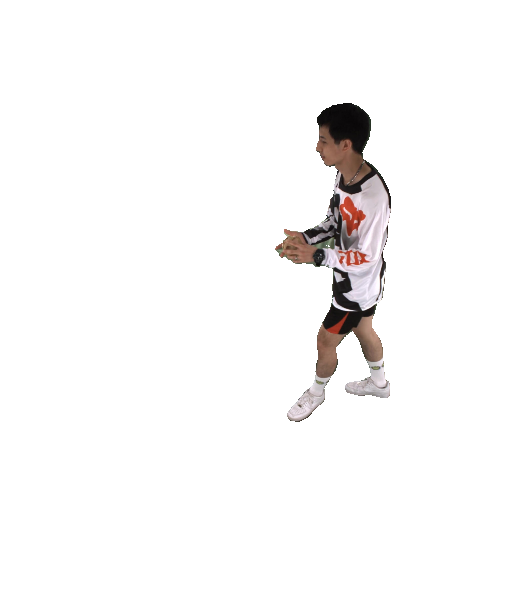}{\zoomx}{\zoomy}{\zoomboxx}{\zoomboxy}{1.2cm}{\zoomlevel}{240 190 80 200}
        \caption*{Ground Truth}
    \end{subfigure}

    \caption{Results on a dynamic scene showing wrongly estimated colors, missing the pattern on the sleeve. Green color artifacts between the arms in the ground truth image are a result of inaccurate foreground masks in the dataset.
    }
    \vspace{-12pt}
    \label{fig:failure}
\end{figure}
\subsection{Ablation Study}

In addition to the comparison to previous approaches, we also conducted an ablation study to analyze the impact of key design choices on the performance of our approach.
We disabled key components of our approach and considered: using only a single camera ("Single cam"), using a single camera and without including masks to our model (“Single cam w/o masks”), using multiple cameras without past video-frames (“w/o temp”), using multiple cameras without Rotary Positional Encodings (“w/o RoPE”), and our full model.
In \Cref{tab:ablation}, we report the results for the whole image and in the inpainted regions, while \Cref{fig:ablation} provides a visual comparison on a challenging scene.
It can be seen that without masks or the positional embedding, our model struggles to have a spatial understanding of the target region, and fails to identify the right color for the arm.
Likewise, without temporal data, the model cannot make use of past frames to retrieve more information about the region.
Interestingly, in this case, the right color is found by the model even without using multiview data, but the full model is still able to produce fewer gray artifacts.

\subsection{Failure Cases}

We observed that fast moving content in the scene can reduce the output quality of our method when the assumption in \Cref{eq:spatiotemporal_coords} no longer holds, \ie that screen-space patch coordinates from \emph{past} frames can be reprojected into the target view using the \emph{current} geometry proxy.
\Cref{fig:failure} shows an example where the region between their arms is mostly occluded, and the fast motion likely causes our model to not relate the missing region to the sleeves of the clothes in other frames.
Additionally, due to slight inaccuracies in the foreground mask of the subject, the green screen background becomes visible in the small gap between the arms in the ground-truth image.
This adds another difficulty to inferring the correct colors in such regions.

\section{Conclusion}

In this work, we introduced a transformer-based, multi-view-aware inpainting method specifically designed for real-time 3D streaming in sparse multi-camera environments.
Our approach functions as a standalone post-processing module, independent of the underlying scene representation, using a novel, spatio-temporal encoding for enhanced feature propagation and a top-k filtering to achieve real-time performance.
Comprehensive evaluations demonstrate that our model outperforms state-of-the-art inpainting methods under real-time constraints, achieving the best trade-off between visual quality and efficiency.

\section*{Acknowledgements}
This work was supported by the European Regional Development Fund (ERDF) and the State of North Rhine-Westphalia as part of the operational program EFRE/JTF-Programm NRW 2021-2027. The project, titled ``Gen-AIvatar'', was funded under the NEXT.IN.NRW competition with the grant agreement No. EFRE-20801085.

Additionally, it has been funded by the Federal Ministry of Education and Research of Germany and the state of North-Rhine Westphalia as part of the Lamarr-Institute for Machine Learning and Artificial Intelligence and by the Federal Ministry of Education and Research under Grant No. 01IS22094A WEST-AI.

The work has also been funded by the Ministry of Culture and Science North Rhine-Westphalia under grant number PB22-063A (InVirtuo 4.0: Experimental Research in Virtual Environments), and by the state of North Rhine Westphalia as part of the Excellency Start-up Center.NRW
(U-BO-GROW) under grant number 03ESCNW18B.
{
    \small
    \bibliographystyle{ieeenat_fullname}
    \bibliography{main}
}

\end{document}